\documentclass[acmtog]{acmart}
\acmSubmissionID{1520}
\usepackage{booktabs} 
\usepackage{color}
\usepackage{makecell}
\usepackage{natbib}
\usepackage{multirow}

\AtBeginDocument{%
  \providecommand\BibTeX{{%
    \normalfont B\kern-0.5em{\scshape i\kern-0.25em b}\kern-0.8em\TeX}}}

\copyrightyear{2025} 
\acmYear{2025} 
\setcopyright{cc}
\setcctype{by-nc-sa}
\setcctype{by-nc-sa}
\acmConference[SA Conference Papers '25]{SIGGRAPH Asia 2025 Conference
 Papers}{December 15--18, 2025}{Hong Kong, Hong Kong}
\acmBooktitle{SIGGRAPH Asia 2025 Conference Papers (SA Conference Papers
 '25), December 15--18, 2025, Hong Kong, Hong
 Kong}
\acmDOI{10.1145/3757377.3763887}
\acmISBN{979-8-4007-2137-3/2025/12}

\begin{document}

\title{LSF-Animation: Label-Free Speech-Driven Facial Animation via Implicit Feature Representation}

\author{Xin Lu}
\authornote{These authors contributed equally to this work.}
\affiliation{%
  \institution{University of Chinese Academy of Sciences}
  \city{Beijing}
  \country{China}
}
\affiliation{%
  \institution{Zhongguancun Academy}
  \city{Beijing}
  \country{China}
}
\email{luxin22@mails.ucas.ac.cn}

\author{Chuanqing Zhuang}
\authornotemark[1]
\affiliation{%
  \institution{University of Chinese Academy of Sciences}
  \city{Beijing}
  \country{China}
}
\email{zhuangchuanqing19@ucas.ac.cn}

\author{Chenxi Jin}
\affiliation{%
  \institution{National University of Singapore}
  \city{Singapore}
  \country{Singapore}
}
\email{jinchenxi@u.nus.edu}

\author{Zhengda Lu}
\affiliation{%
  \institution{University of Chinese Academy of Sciences}
  \city{Beijing}
  \country{China}
}
\email{luzhengda@ucas.ac.cn}

\author{Yiqun Wang}
\authornote{These authors are co-corresponding authors.}
\affiliation{%
  \institution{Chongqing University}
  \city{Chongqing}
  \country{China}
}
\email{yiqun.wang@cqu.edu.cn}

\author{Wu Liu}
\affiliation{%
  \institution{University of Science and Technology of China}
  \city{Hefei}
  \country{China}
}
\email{liuwu@ustc.edu.cn}

\author{Jun Xiao}
\authornotemark[2]
\affiliation{%
  \institution{University of Chinese Academy of Sciences}
  \city{Beijing}
  \country{China}
}
\affiliation{%
  \institution{Zhongguancun Academy}
  \city{Beijing}
  \country{China}
}
\email{xiaojun@ucas.ac.cn}









\begin{abstract}
Speech-driven 3D facial animation has attracted increasing interest since its potential to generate expressive and temporally synchronized digital humans. While recent works have begun to explore emotion-aware animation, they still depend on explicit one-hot encodings to represent identity and emotion with given emotion and identity labels, which limits their ability to generalize to unseen speakers. Moreover, the emotional cues inherently present in speech are often neglected, limiting the naturalness and adaptability of generated animations.
In this work, we propose LSF-Animation,
a novel framework that eliminates the reliance on explicit emotion and identity feature representations. Specifically, LSF-Animation implicitly extracts emotion information from speech and captures the identity features from a neutral facial mesh, enabling improved generalization 
to unseen speakers and emotional states without requiring manual labels.
Furthermore, we introduce a Hierarchical Interaction Fusion Block (HIFB), which employs a fusion token to integrate dual transformer features and more effectively integrate emotional, motion-related and identity-related cues. Extensive experiments conducted on the 3DMEAD dataset demonstrate that our method surpasses recent state-of-the-art approaches in terms of emotional expressiveness, identity generalization, and animation realism.
The source code will be released at: https://github.com/Dogter521/LSF-Animation.

\end{abstract}



\maketitle

\begin{figure}[h]
  \includegraphics[width=\linewidth]{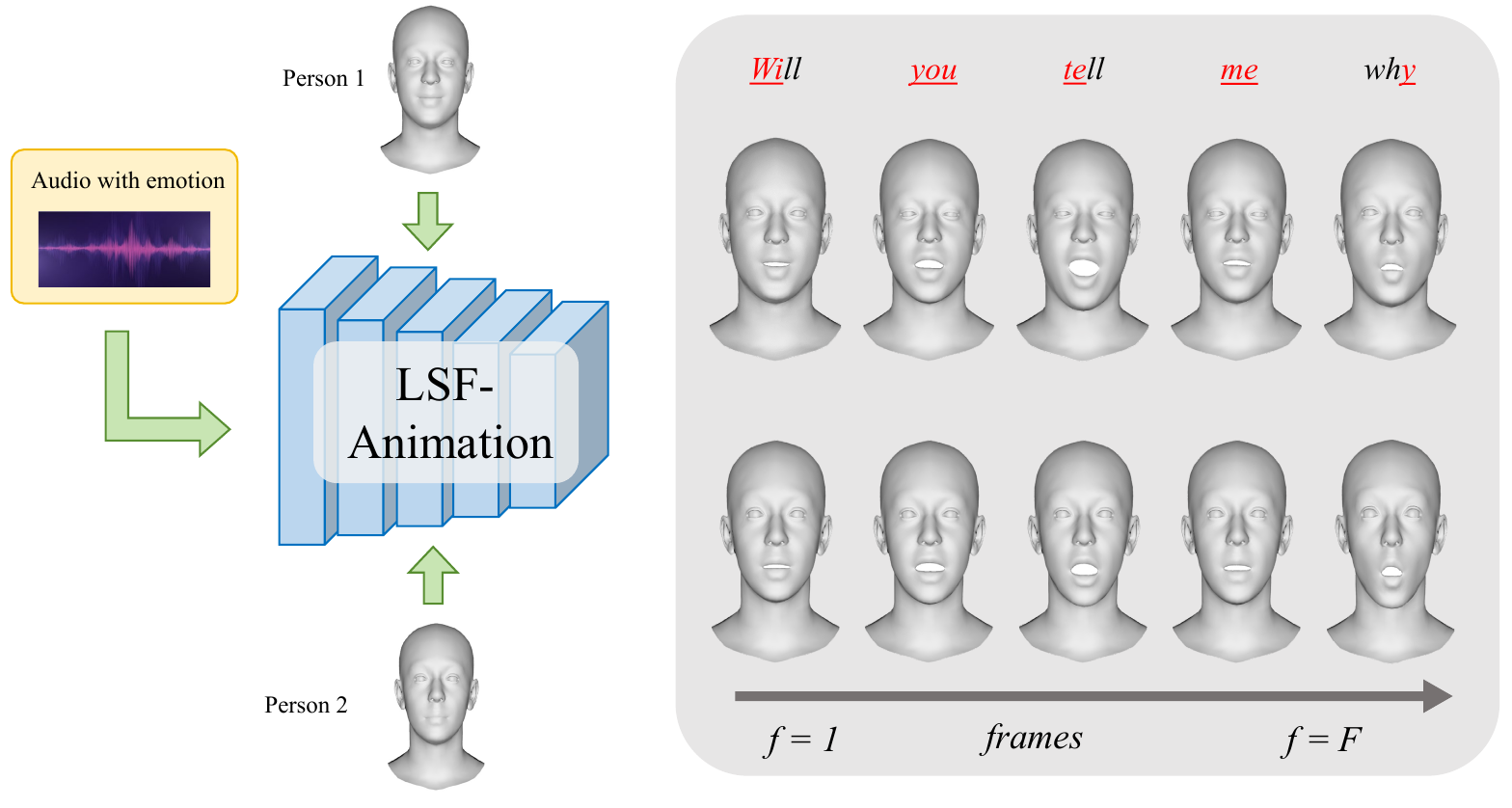}
  \caption{ \textbf{LSF-Animation} generates expressive 3D facial animations in a fully label-free manner. Given emotional audio and a neutral face mesh, it extracts dynamic emotion cues and identity representations directly from the inputs, without relying on predefined labels. The results demonstrate accurate lip synchronization, faithful reflection of the emotional content in speech, and strong generalization to unseen identities.
}

  \Description{.}
  \label{fig:sample}
\end{figure}

\section{Introduction}
In recent years, 3D facial animation has shown meaningful application potential across film visual effects, game character development, and Extended reality (XR) digital humans. However, traditional production pipelines heavily rely on costly motion capture systems and manual adjustments by skilled artists, significantly limiting scalability and efficiency. To address these challenges, researchers have increasingly focused on automated speech-driven methods~\cite{VOCA,karras2017audio,facediffuser,codetalker}, aiming to achieve natural synchronization between speech content and facial dynamics for immersive interactions.

Early approaches to 3D facial animation primarily relied on rule-based procedural techniques~\cite{charalambous2019audio,jali}, which mapped audio signals to animation through hand-crafted linguistic rules. Although intuitive and accessible to artists, these methods were restricted to basic lip synchronization and lacked the capacity to generate facial emotional dynamics. With the development of deep learning, researchers have proposed the end-to-end neural models~\cite{faceformer,facexhubert,karras2017audio,meshtalk,taylor2017deep,imitator}, enabling more accurate and data-driven facial motion synthesis directly from speech. However, these models typically train on small-scale 3D vertex-based datasets (e.g., VOCASET~\cite{VOCA}, Multiface~\cite{multiface}, BIWI~\cite{fanelli20103}) with limited speaker identities and emotional expressions, resulting in constrained generalization and expressiveness.

To address the lack of realism, recent works have explored non-deterministic generation frameworks~\cite{facetalk,ng2022learning,facediffuser,sun2024diffposetalk,yang2024probabilistic}, which leverage probabilistic modeling to synthesize multiple plausible animation sequences from the same audio. Meanwhile, 
researchers have also studied generation methods based on rigged characters and blend shapes~\cite{villanueva2022voice2face,park2023said,facediffuser}, as well as holistic approaches that jointly animate face and body~\cite{liu2023emage,ng2024audio}. 
Nevertheless, most existing models have limited ability in fine-grain emotion control, primarily due to the lack of richly annotated datasets and the reliance on rigid emotion encodings.

Several recent studies have attempted to address emotional modeling more explicitly. Media2Face~\cite{media2face} introduced a high-quality 4D dataset with categorical emotion labels, and EmoTalk~\cite{emotalk} and EMOTE~\cite{EMOTE} proposed pipelines based on reconstructed 3D datasets to incorporate emotion into speech-driven animation. 
Although these methods have improved the ability of emotion expression to some extent, they still heavily rely on predefined emotion labels, which limits their adaptability and generalization capabilities.
ProbTalk3D~\cite{probtalk3d} introduced a probabilistic two-stage framework that integrates emotional control and diverse output generation, yet it depends on explicit one-hot labels for emotion and identity, constraining its scalability to the label-free applications in real-world.

To address these limitations, we propose LSF-Animation: label-free speech-driven facial animation via implicit feature representation, a novel framework that eliminates the reliance on explicit emotion and identity feature representations.
Specifically, LSF-Animation implicitly extracts emotion information from speech and captures the identity features from a neutral facial mesh, instead of adjusting the model on known emotion and identity inputs. This design allows for flexible, data-driven generation of personalized facial animations and significantly enhances generalization to unseen speakers and emotional states without requiring manual labels.
Furthermore, we introduce a HIFB module to effectively integrate emotional and motion information embedded within speech. This module includes a fusion token to facilitate dense hierarchical interactions between two transformer branches used for feature extraction. This mechanism notably enhances the stability and expressiveness of upper-face animation, a particularly challenging aspect of facial dynamics generation.
Our contributions are summarized as follows:

\begin{itemize}
\item[$\bullet$] We propose LSF-Animation, a speech-driven 3D facial animation representation that eliminates the reliance on one-hot emotion and identity labels. The model extracts expressive emotional cues directly from speech, enabling fully label-free inference, and captures identity features from a neutral face mesh, enhancing generalization to unseen speakers.

\item[$\bullet$] We design a HIFB module to effectively integrate emotional and motion cues from speech, notably improving expressive facial dynamics, particularly in upper-face stability.

\item[$\bullet$]We conduct comprehensive experiments on the 3DMEAD dataset, demonstrating that our model outperforms recent state-of-the-art methods in emotional expressiveness, identity generalization, and animation realism through quantitative evaluation, qualitative analysis, and perceptual user studies.

\end{itemize}

\section{RELATED WORK}
This section reviews recent advances in deep learning-based 3D facial animation synthesis, with a focus on speech-driven generation and emotion-controllable modeling. While our emphasis is on 3D animation, it is worth noting that a large body of research has focused on 2D facial animation synthesis using deep learning techniques ~\cite{ji2022eamm,stypulkowski2024diffused,zhang2023sadtalker}, including keypoint-driven and GAN-based methods. These 2D approaches are widely applied in virtual avatars and online communication but differ fundamentally in their representation space, geometric constraints, and application scope compared to 3D methods. Therefore, we do not discuss them further in this paper. Similarly, several studies have addressed facial representation learning, tracking, and 3D reconstruction ~\cite{danvevcek2022emoca,egger20203d,giebenhain2024mononphm,flame}, which serve as important upstream tasks in the animation pipeline but fall outside the generation-oriented focus of this review. We categorize the literature into two primary streams: speech-driven 3D facial animation and emotion-aware facial animation. Within each, we highlight representative efforts that explore non-deterministic generation and emotional conditioning.

\subsection{Audio driven 3D Facial Animation}
Contemporary speech-driven 3D facial animation approaches can be broadly divided into phoneme-based and data-driven paradigms. Phoneme-based methods, such as JALI~\cite{jali}, are compatible with standard animation pipelines but depend on intermediate phoneme inputs to model co-articulation effects~\cite{charalambous2019audio}. In contrast, data-driven methods like VOCA~\cite{VOCA} use CNNs to directly map raw audio to 3D meshes, introducing one-hot identity vectors for identity-specific style. Mesh-Talk~\cite{meshtalk} explores latent space disentanglement between speech and facial animation but faces efficiency challenges due to limited audio context and high computational cost.

Transformer-based models mark a significant evolution. The first method to adopt an autoregressive transformer backbone was FaceFormer~\cite{faceformer}, which leverages attention for temporal audio feature extraction. EmoTalk~\cite{emotalk} extends this with a cross-attention module for explicit emotion conditioning. FaceDiffuser~\cite{facediffuser} applies diffusion modeling but lacks identity-specific generation. To address this, DiffusionTalker~\cite{chen2023diffusiontalker} introduces contrastive learning to embed identity-aware features, achieving stylized and consistent facial animations.
These works collectively shift modeling from blendshape or parameter space to full 3D vertex-based representations~\cite{zhou2018visemenet,bao2023learning}. However, most models depend on one-hot identity encodings, which limits generalization. Imitator~\cite{imitator} proposes a two-stage adaptation strategy using reference videos, improving cross-identity generalization but requiring additional reference material and suffering from performance degradation with limited inputs.
Mimic~\cite{mimic} overcomes these limitations by learning disentangled latent spaces for speaking style and semantic content. It employs a combination of a style classifier, inverse classifier, content contrastive loss, and cycle consistency to synthesize expressive, identity-adaptive animations without needing per-subject fine-tuning.

Despite these advances, emotion modeling remains underdeveloped. Most approaches focus on lip-speech alignment, yet fail to capture the emotional nuances in audio, reducing realism and limiting user engagement in emotionally driven applications.

\begin{figure*}
  \includegraphics[width=0.7\textwidth]{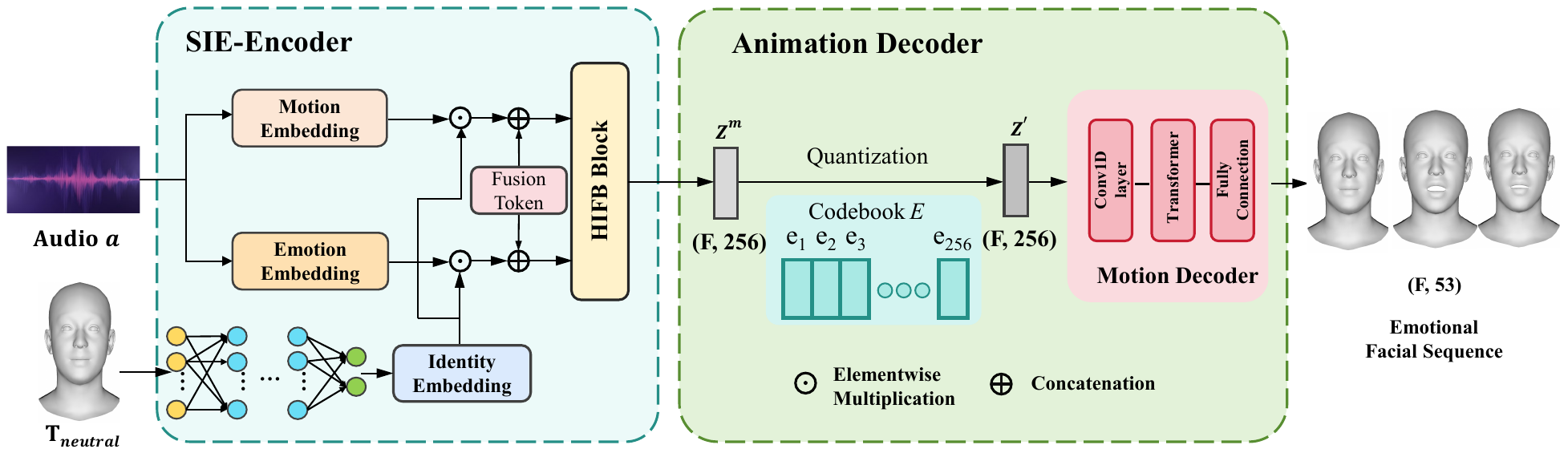}
  \caption{ The overview of our LSF-Animation framework. }

  \Description{.}
  \label{fig:LSF-Animation}
\end{figure*}

\subsection{Emotion-Controllable 3D Facial Animation}
Early work on emotion-controllable facial animation focused on directly synthesizing 4D expressions from labels~\cite{chang2018expnet,chang2017faceposenet,Gecer_2019}, but dense vertex data posed challenges for capturing fine emotional details. As facial landmark detection improved~\cite{detection0,detection1}, researchers adopted trajectory-based methods. For example, Motion3D~\cite{S2D} utilizes WGANs to generate the Square Root Velocity Function (SRVF) of facial landmarks, which is decoded into mesh displacements via a sequence-to-deformation (S2D) module.
More recent methods generate full expression sequences from neutral landmarks~\cite{lu2023landmark,lu2024fc,4DFM,VAE_trans}, yet often produce fixed-length animations and exhibit poor scalability across varying identities and emotional states. 
In addition, recent works~\cite{rebuttal_cite_1,rebuttal_cite_2} attempt to disentangle emotion and content directly from audio. Specifically,~\cite{rebuttal_cite_1} learns two separate latent audio spaces for emotion and speech content, while~\cite{rebuttal_cite_2} extracts emotion dynamics from source videos and applies them to audio-driven facial movements using unsupervised motion representations. These approaches, however, typically rely on paired audio clips with identical content but different emotions, which are difficult to obtain in real-world scenes.

To integrate emotion in speech-driven settings, Karras et al.~\cite{karras2017audio} incorporated implicit emotion cues using CNNs trained on small datasets. EmoTalk~\cite{emotalk} enhances this with an emotion disentanglement encoder and attention-based decoder, using a combined dataset of RAVDESS and HDTF. 
Besides, EMOTE~\cite{EMOTE} extends this direction by introducing the 3DMEAD dataset with rich emotional diversity and applying a temporal VAE with a non-autoregressive transformer for real-time expressive animation. However, it still yields deterministic outputs.
Media2Face~\cite{media2face} proposes a two-stage diffusion framework, using VAE-based geometry encoding and CLIP-driven emotional control, but lacks publicly available data and metrics to evaluate diversity. Emo-VOCA~\cite{nocentini2025emovoca} addresses dataset limitations by generating a synthetic corpus combining neutral speech-driven motion with expressive variants. Its generator takes audio, emotion category, and intensity as inputs.
ProbTalk3D~\cite{probtalk3d} further integrates non-deterministic modeling and emotion control through a two-stage VQ-VAE pipeline built on 3DMEAD, learning a latent space of emotional motion. It outperforms prior work in perceptual and objective evaluations. Nevertheless, it still depends on explicit emotion and identity labels, restricting deployment in open-world scenarios.

In summary, although significant progress has been made in both speech-driven and emotion-aware 3D facial animation, current methods often struggle to balance emotional fidelity, identity generalization, and generative diversity. 
In particular, many existing approaches still rely on paired data with the same content but different emotions, which is difficult to obtain in real-world scenarios and limits practical deployment.
These remain key challenges that our proposed representation aims to address.

\section{Methodology}\label{method}
In this section, we present the details of our proposed methodology. We begin by introducing the framework overview, followed by a comprehensive description of our Implicit Feature Representation of LSF-Animation, which eliminates the need for one-hot encodings of emotion and identity by leveraging implicit representations extracted from speech and a neutral facial mesh. We then describe the architecture of the HIFB, which enables fine-grained integration of emotional and motion features through a fusion mechanism. Finally, we introduce our training strategy for our whole framework.

\subsection{Framework Overview}
Our framework consists of two main components: a \textit{Speech-Aware Identity-Emotion Encoder (SIE-Encoder)} and an \textit{Animation Decoder}, as illustrated in Fig.~\ref{fig:LSF-Animation}. The goal is to generate identity-generalizable and emotion-aware 3D facial animations from speech in a fully label-free manner.

The SIE-Encoder is responsible for extracting implicit emotional and identity representations from the input speech and neutral face mesh. It consists of two submodules: (1) the Implicit Feature Representation, which extracts motion and emotion features using from input audio \(a\) while extracting identity features from the neutral face mesh $ T_{\text{neutral}}$; and (2) the \textit{HIFB} module, which fuses the extracted motion, emotion, and neutral-face-based identity features.

The output of the SIE-Encoder is projected into a discrete latent space of facial motion, learned through a VQ-VAE-based Animation Decoder. This latent space is parameterized by a codebook \(E = \{ e_k \in \mathbb{R}^C \}_{k=1}^N\), where each code corresponds to a representative facial motion pattern. By quantizing the fused feature representation to its nearest codebook entry, the model learns to represent complex facial dynamics in a compact and structured manner. The decoder then reconstructs frame-wise 3D facial animation sequences from these latent codes. Each frame is represented using a 53-dimensional FLAME parameter vector (50 expression coefficients + 3 jaw rotations), following the 3DMEAD~\cite{EMOTE} dataset.

\subsection{Implicit Feature Representation}\label{section:IFP}

In this section, we introduce the Implicit Feature Representation of our proposed LSF-Animation, a novel representation for generating expressive 3D facial animations with an audio and a neutral face mesh, as illustrated in Fig.~\ref{fig:LSF-Animation}.
It extracts implicit emotional and identity embedding from the input audio and neutral face mesh, thereby it eliminates the need for prior explicit emotion categories and identity instance labels (e.g., one-hot encoding), enabling label-free generation and improving generalization to unseen speakers.

\paragraph{Speech-based Emotion Embedding.}
Audio sequences contain rich and subtle emotional expressions and facial motion. Previous works have used the HuBERT model to extract audio features or employed one-hot encoding of explicit labels to provide emotion category priors. However, they have not achieved a high degree of coupling between audio, emotion, and facial motion.
Inspired by the task of audio-based emotion recognition, we propose to extract emotion embedding with learned continuous representations from speech and replace the traditional one-hot emotion labels to effectively capture fine-grained affective signals.

We build our speech-based emotion embedding module on top of the pretrained Emotion2vec model, a self-supervised framework trained on a large-scale emotional speech dataset. Given a raw waveform \(a\) sampled at 16 kHz, Emotion2vec processes the input through convolutional and transformer layers to extract high-level frame-wise emotional embeddings. 
Unlike common approaches that rely on sequence-level classification or generate pseudo-labels through emotion recognition tasks, our representation leverages the continuous frame-level features directly as emotional input. In our setting, these embeddings are sampled at 50Hz, with each frame represented by a 768-dimensional vector, offering fine-grained temporal resolution. This design enables our model to preserve detailed affective variations over time and avoids the need for intermediate emotion classification steps.

Formally, the emotional embedding sequence is defined as:
\[
e_{1:T} = E(a), \quad e_{1:T} \in \mathbb{R}^{T \times d_e}
\tag{1}
\]
where \(E\) denotes the speech-based emotion embedding model, \(a\) is the raw audio input, \(T\) is the number of frames, and \(d_e = 768\) is the feature dimension. This formulation enables our representation to generate expressive and emotion-consistent animations in a label-free manner, allowing for flexible adaptation to diverse emotional states (e.g., coarticulated variations in emotion intensity).

\paragraph{Neutral-face-based Identity Embedding.}
Previous work FC-4DFS~\cite{lu2024fc} has shown that the identity information for facial animation can be obtained from neutral facial landmarks, without the need for additional identity labels.
In this work, we design a lightweight neutral-face-based identity embedding module to efficiently extract speaker-specific identity feature embeddings from neutral FLAME shape parameters, promoting generalization of speech-driven facial animation synthesis to unseen identities.

Specifically, the extractor takes the FLAME shape parameter \(T_{\text{neutral}} \in \mathbb{R}^{300}\) as input, which serves as a compact and semantically meaningful representation of the subject's neutral 3D facial structure. Compared to directly using the high-dimensional neutral mesh, this parameter-based representation is more structured, lower-dimensional, and easier to process, providing a lightweight alternative for identity encoding.
Subsequently, we use a lightweight MLP model $F_{\text{id}}$ to extract neutral identity features $z_{\text{id}}$, avoiding the computational overhead of spiral convolutions used in FC-4DFS, as shown in the following equation
\[
z_{\text{id}} = F_{\text{id}}(T_{\text{neutral}}), \quad z_{\text{id}} \in \mathbb{R}^{d_{\text{id}}}
\tag{2}
\]

By jointly leveraging emotional embeddings extracted from audio and identity features captured from neutral face geometry, our LSF-Animation enables the synthesis of expressive and identity-consistent facial animations without relying on any manually annotated emotion or identity labels.
The effectiveness and generalization capability of our representation are validated through comprehensive quantitative evaluations presented in Section~\ref{section:Quantitative}.

\begin{figure}
  \includegraphics[width=0.8\linewidth]{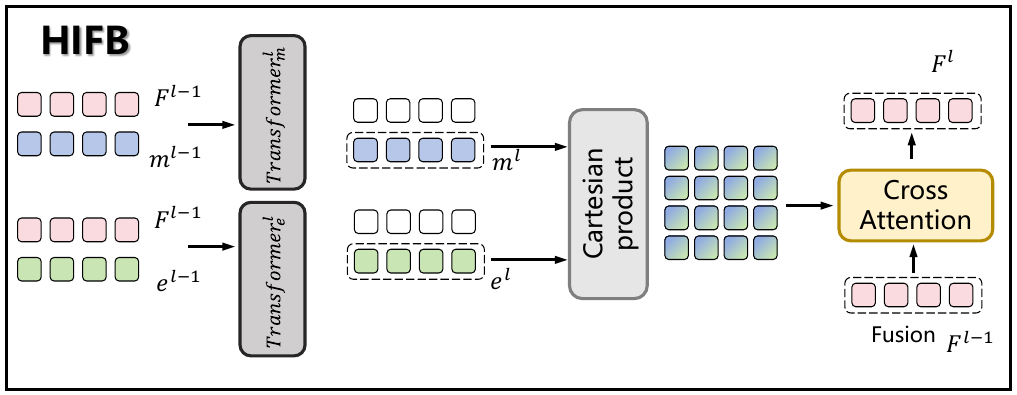}
  \caption{ The framework of our Hierarchical Interaction Fusion Block (HIFB).
}

  \Description{.}
  \label{fig:HIFB}
\end{figure}

\subsection{Hierarchical Interaction Fusion Block (HIFB)}

Although the speech-based emotion embedding and neutral-face-based identity embedding modules extract implicit feature representations of emotion and identity with expressive capabilities, integrating and aligning them with motion features remains a crucial step in generating detailed facial animations.
Previous works, such as ProbTalk3D, construct a global sequence-level style vector directly from emotion and identity information, which discards fine-grained temporal details and weakens the expressiveness and richness of facial animation. Moreover, traditional feature fusion methods~\cite{lu2024fc, faceformer} are difficult to achieve temporal alignment between emotion and motion sequences.

To overcome this limitation, we propose the \textit{HIFB} module, a task-specific cross-modal module that tightly integrates emotional cues \(e_{1:T} \in \mathbb{R}^{T \times d_e}\) and motion features \(m_{1:T} \in \mathbb{R}^{T \times d_m}\), both extracted from the same speech signal via Emotion2vec and HuBERT, respectively, as illustrated in Fig.~\ref{fig:HIFB}. Rather than relying on discrete labels, HIFB directly operates on raw frame-level features, preserving the temporal structure critical to facial dynamics.

We begin by projecting both emotion and motion features into a shared latent space and modulating them with identity information extracted from the neutral face:
\begin{equation}
\quad s_{\text{id}} = W_{\text{style}} z_{\text{id}},\tilde{m}_{1:T} = W_m m_{1:T} \odot s_{\text{id}}, \quad \tilde{e}_{1:T} = W_e e_{1:T} \odot s_{\text{id}}
\tag{3}
\end{equation}
where \(z_{\text{id}}\) is the identity code, and \(\odot\) denotes element-wise multiplication.

To enable progressive alignment across modalities, we initialize a set of learnable fusion tokens \(\mathbf{F}_0 \in \mathbb{R}^{n_f \times d}\), inspired by audio-visual joint training in masked autoencoders (MAE)~\cite{deepavfusion}. These fusion tokens are updated over \(L\) layers based on bidirectional interactions with motion and emotion streams. In our implementation, \(d_e = d_m = d = 768\), ensuring unified dimensionality across streams.

At each fusion layer \(l\), the inputs from each modality are separately concatenated with the current fusion tokens and passed through their corresponding Transformer blocks. The updated frame-level representations are then obtained by removing the concatenated tokens:
\begin{equation}
\begin{aligned}
\tilde{m}_{1:T}^l &= \text{Transformer}_m^l([\mathbf{F}^{l-1}; \tilde{m}^{l-1}]), \\
\tilde{e}_{1:T}^l &= \text{Transformer}_e^l([\mathbf{F}^{l-1}; \tilde{e}^{l-1}])
\end{aligned}
\tag{4}
\end{equation}

After updating the streams, we compute the Cartesian product of all pairwise combinations of motion and emotion features and use them to update the fusion tokens through cross-attention:
\begin{equation}
\mathbf{F}^l = \text{CrossAttn}\left(\mathbf{F}^{l-1}, \left\{ \langle \tilde{m}_i^{l}, \tilde{e}_j^{l} \rangle \right\}_{i,j=1}^T \right)
\tag{5}
\end{equation}
where $\langle \cdot, \cdot \rangle$ denotes the Cartesian product notation, which here refers to concatenating motion and emotion features across the feature dimension for each pair of temporal indices $(i, j)$.
This densely constructed interaction space enables the fusion tokens to capture rich temporal cross-modal dependencies.

After \(L\) layers of hierarchical fusion, we take the final motion stream as output:
\[
m_{\text{final}} = \tilde{m}^L_{1:T}
\]
This progressive and bidirectional fusion architecture allows the model to preserve fine-grained emotional variation and improves upper-face expressiveness, identity consistency, and overall realism in the generated animation.

\subsection{Training Strategy}

We train our model in two stages to learn expressive and identity-aware 3D facial animations from audio in a fully label-free manner.

In the first stage, we train a VQ-VAE-based motion autoencoder to model the distribution of facial motion, following prior work~\cite{probtalk3d}. We learn a discrete codebook \( Z = \{ z_k \in \mathbb{R}^C \}_{k=1}^N \), which serves as a latent space to represent prototypical facial motion patterns. Given an input motion sequence \(X_{1:T}\), the encoder \(E\) maps it to latent features \(\hat{Z}\), which are quantized to their nearest codebook entries:
\[
Z_q = Q(\hat{Z}) := \mathop{\arg\min}_{z_k \in Z} \| \hat{z}_t - z_k \|_2.
\]
The decoder \(D\) reconstructs the facial motion sequence from these quantized latent codes:
$X_{1:T} = D(Z_q) = D(Q(E(X_{1:T}))).$
This learned latent space captures structured motion priors and is used in the next stage for speech-to-motion generation.

In the second stage, we train the SIE-Encoder to predict latent motion representations from raw audio \(a\) and a neutral face mesh \(T_{\text{neutral}}\). The SIE-Encoder \(M\) extracts emotion, motion, and identity features and fuses them through the HIFB module. The fused features are projected into the VQ-VAE latent space and decoded into a sequence of 3D facial motion parameters:
\[
\hat{X}_{1:F} = D(M(a, T_{\text{neutral}}))
\tag{2}
\]
Here, the neutral face mesh serves as an identity reference and does not carry any expression. This two-stage design allows the model to decouple motion generation from identity and emotion representation learning, thereby enhancing generalization to unseen speakers and emotional variations.

\begin{table*}[t]
\centering
\caption{Comparison of quantitative results with SOTA methods. $\downarrow$ means lower is better, $\uparrow$ means higher is better. All metrics are in mm.}
\label{tab:quant_results}
\resizebox{\textwidth}{!}{
\begin{tabular}{lcccccc}
\toprule
\textbf{Model} & 
\begin{tabular}{c}
\textbf{MVE} $\downarrow$ \\
($\times10^{-3}$ mm)
\end{tabular} &
\begin{tabular}{c}
\textbf{LVE} $\downarrow$ \\
($\times10^{-4}$ mm)
\end{tabular} &
\begin{tabular}{c}
\textbf{FDD} $\downarrow$ \\
($\times10^{-6}$ mm)
\end{tabular} &
\begin{tabular}{c}
\textbf{MEE} $\downarrow$ \\
($\times10^{-4}$ mm)
\end{tabular} &
\begin{tabular}{c}
\textbf{CE} $\downarrow$ \\
($\times10^{-5}$ mm)
\end{tabular} &
\begin{tabular}{c}
\textbf{Diversity} $\uparrow$ \\
($\times10^{-3}$ mm)
\end{tabular} \\
\midrule
FaceFormer~\cite{faceformer}            & 2.6139 & 2.3471 & 1.6145 & --     & --     & -- \\
CodeTalker~\cite{codetalker}            & 2.2207 & 2.1045 & 3.0540 & --     & --     & -- \\
EMOTE~\cite{EMOTE}  & 	1.3395 & 1.2936 & 0.7327 & 1.2101 & 1.0544 & -- \\
FaceDiffuser (DDPM)~\cite{facediffuser}   & 1.6762 & 1.3463 & 1.7015 & 1.3463 & 1.3462 & 0.0005 \\
FaceDiffuser (DDIM)    & 1.5901 & \underline{1.1315} & 0.8243 & \underline{1.0615} & \underline{1.0610} & 0.0144 \\

ProbTalk3D~\cite{probtalk3d}            & \underline{1.2933} & 1.2708 & \underline{0.4845} & 1.1852 & 1.0691 & \textbf{0.4310} \\
\textbf{LSF-Animation (Ours)} & \textbf{1.2244} & \textbf{1.0985} & \textbf{0.4724} & \textbf{1.0177} & \textbf{0.9225} & \underline{0.4223} \\
\bottomrule
\end{tabular}
}

\end{table*}

\section{EXPERIMENTS AND RESULTS}\label{section:results}
\subsection{ Experimental details}\label{section:details}

\paragraph{Datasets Selection} 
Several recent datasets have been introduced to support speech-driven 3D facial animation with emotional content. M2F-D~\cite{media2face} offers a high-quality 4D corpus with annotated emotion labels, but the lack of public access to the dataset and code limits reproducibility and fair benchmarking. EmoVOCA~\cite{nocentini2025emovoca} constructs a synthetic dataset by combining neutral speech-driven motion with separately recorded expressive mesh sequences. While enabling emotion-conditioned generation, the use of neutral audio weakens audiovisual coherence due to the absence of emotional prosody.

In contrast, we adopt the 3DMEAD dataset~\cite{EMOTE}, which provides synchronized 3D facial mesh sequences aligned with emotionally expressive speech. It is reconstructed from the 2D MEAD dataset~\cite{wang2020mead} using DECA~\cite{deca} and MICA~\cite{mica}, following the pipeline in EMOTE~\cite{EMOTE}. Compared to synthetic or loosely aligned datasets, 3DMEAD ensures natural temporal alignment between facial motion and emotional speech, critical for realistic and expressive animation.

3DMEAD contains 47 speaker identities recorded under eight emotion categories—\textit{neutral, happy, sad, surprised, fear, disgusted, angry}, and \textit{contempt}. Each non-neutral emotion includes three intensity levels (\textit{weak, medium, strong}) with 30 short sentences per level, and 40 neutral sentences per subject. Mesh sequences are sampled at 25 fps, making 3DMEAD a large-scale, emotionally rich dataset well-suited for training and evaluating speech-driven animation models.

\paragraph{Dataset Split.} 
Prior works have adopted varying data split strategies, each emphasizing different evaluation priorities. 
For instance, ProbTalk3D uses a sentence-level split, where all identities share the same set of sentences across the training, validation, and test sets. While this design facilitates consistent evaluation on sentence content, it prevents assessment of a model’s ability to generalize to unseen speaker identities, since every identity is present in all subsets.
Following EMOTE~\cite{EMOTE}, we adopt a subject-level split strategy, where each speaker identity appears in only one of the training, validation, or test sets. This approach enables explicit evaluation of identity generalization to unseen speakers. To ensure robust quantitative evaluation, we also reserve a held-out test set with full ground-truth annotations. Although subject-level splitting reduces the overall amount of training data compared to sentence-level splits, the large scale and emotional richness of the 3DMEAD dataset still provide sufficient coverage for effective learning and perceptual fidelity.

\paragraph{Implement Details} 
We implement our proposed model using the PyTorch Lightning framework. All experiments are conducted on a workstation equipped with an NVIDIA RTX 4090 GPU and an Intel(R) Xeon(R) Platinum 8368 CPU @ 2.40GHz.
In stage 1, we train the motion autoencoder using the AdamW optimizer with a learning rate of \(1 \times 10^{-4}\). In stage 2, the prediction model is optimized using Adam with a learning rate of \(1 \times 10^{-5}\). For both stages, we apply early stopping to prevent overfitting: if the validation loss does not improve for 5 consecutive epochs, training is terminated and the best model weights are retained.
Stage 1 converges after 21 epochs (approximately 6.5 hours), while stage 2 completes in 36 epochs (approximately 30 hours). Throughout training, we monitor the validation loss to guide model selection and ensure stability.

\subsection{Quantitative Evaluation}\label{section:Quantitative}

\paragraph{Evaluation Metrics}  
We follow ProbTalk3D and evaluate our model using six metrics covering accuracy, expressiveness, and diversity. \textbf{MVE}, \textbf{LVE}, and \textbf{FDD} are computed from a single generated sequence to assess deterministic accuracy—capturing global error, lip-sync quality, and upper-face motion consistency, respectively. In contrast, \textbf{MEE}, \textbf{CE}, and \textbf{Diversity} evaluate non-deterministic outputs across multiple samples, measuring mean prediction accuracy, distributional coverage, and output variability. All metrics are computed in vertex space after FLAME reconstruction to ensure comparability across models.

\paragraph{Comparison Study.} 
As shown in Table~\ref{tab:quant_results}, our proposed method LSF-Animation consistently outperforms prior state-of-the-art models across most evaluation metrics. Compared to deterministic baselines such as FaceFormer, CodeTalker and the 2D-based method EMOTE, LSF-Animation achieves significantly lower MVE and LVE, demonstrating improved accuracy in both global facial motion and lip synchronization. Although FaceDiffuser (DDIM) performs competitively in terms of distribution-level metrics such as MEE and CE, it suffers from high global error (MVE) and low Diversity, indicating limited motion fidelity and poor generative variability. In contrast, LSF-Animation strikes a better balance between accuracy, expressiveness, and diversity, highlighting the effectiveness of our label-free framework.

Among probabilistic methods, ProbTalk3D is currently the strongest non-deterministic generation model in terms of overall performance. It achieves a good balance across accuracy, expressiveness, and diversity. However, our LSF-Animation model further improves upon ProbTalk3D in nearly all metrics: MEE is improved by \textbf{14.1\%} (from 1.1852 to 1.0177), CE by \textbf{13.7\%} (from 1.0691 to 0.9225), MVE by \textbf{5.32\%} (from 1.2933 to 1.2244), LVE by \textbf{13.5\%} (from 1.2708 to 1.0985), and FDD by \textbf{2.50\%} (from 0.4845 to 0.4724).
In addition, we observe a slight decrease in the Diversity metric by \textbf{2.02\%} (from 0.4310 to 0.4223).
This is because our method provides a neutral facial prior, enabling the generated results to maintain diversity while better conforming to the motion patterns of a neutral face. 

These results suggest that LSF-Animation achieves a better trade-off between lip-sync accuracy, upper-face expressiveness, and generative diversity. Moreover, by leveraging label-free emotion and identity embeddings, our model exhibits stronger generalization to unseen speakers, advancing the state of the art in probabilistic facial animation synthesis.

\begin{table*}[t]
\centering
\caption{Ablation study categorized into
representation strategy and fusion strategy. $\downarrow$ means lower is better, $\uparrow$ means higher is better. All metrics are in mm. }
\label{tab:ablation_results}
\resizebox{\textwidth}{!}{%
\begin{tabular}{llcccccc}
\toprule
\textbf{Category} & \textbf{Model Variant} & 
\begin{tabular}{c}
\textbf{MVE} $\downarrow$ \\
($\times10^{-3}$ mm)
\end{tabular} &
\begin{tabular}{c}
\textbf{LVE} $\downarrow$ \\
($\times10^{-4}$ mm)
\end{tabular} &
\begin{tabular}{c}
\textbf{FDD} $\downarrow$ \\
($\times10^{-6}$ mm)
\end{tabular} &
\begin{tabular}{c}
\textbf{MEE} $\downarrow$ \\
($\times10^{-4}$ mm)
\end{tabular} &
\begin{tabular}{c}
\textbf{CE} $\downarrow$ \\
($\times10^{-5}$ mm)
\end{tabular} &
\begin{tabular}{c}
\textbf{Diversity} $\uparrow$ \\
($\times10^{-3}$ mm)
\end{tabular} \\
\midrule
\multirow{6}{*}{\shortstack{Representation \\ Strategy}}
& w/o style one-hot                   & 1.3408 & 1.3274 & 0.5386 & 1.2442 & 1.1145 & \underline{0.4146}\\
& w/o iden                   & 1.5364 & 1.9328 & 	1.0716 & 1.7759 & 1.5605 & 0.3279 \\
& Emo one-hot + shape parameter & 1.3555 & 1.3202 & 	0.5057 & 1.2379 & 	1.1128 & 0.4087\\
& Baseline (gt one-hot)              & 1.2933 & 1.2708 & \textbf{0.4845} & 1.1852 & 1.0691 & \textbf{0.4310} \\
& + Emotion Feature                  & \underline{1.2765} & \underline{1.2184} & 0.6278 & \underline{1.1516} & \underline{1.0524} & 0.3944 \\
& + Emotion Feature + Iden Feature   & \textbf{1.2380} & \textbf{1.1504} & \underline{0.4867} & \textbf{1.0828} & \textbf{0.9753} & 0.3909 \\
\midrule
\multirow{3}{*}{Fusion Strategy} 
& Gate fusion                        & 1.2662 & 1.1315 & \underline{0.7671} & \underline{1.0281} & 9.3436 & 0.4216 \\
& Cross-attention fusion            & \textbf{1.1901} & \underline{1.1312} & 0.9337 & 1.0388 & \underline{0.9247} & \textbf{0.4458} \\
& \textbf{LSF-Animation (HIFB)}    & \underline{1.2244} & \textbf{1.0985} & \textbf{0.4724} & \textbf{1.0177} & \textbf{0.9225} & \underline{0.4223} \\
\bottomrule
\end{tabular}}
\end{table*}

\subsection{Qualitative Evaluation}
\begin{figure*}
  \includegraphics[width=0.8\linewidth]{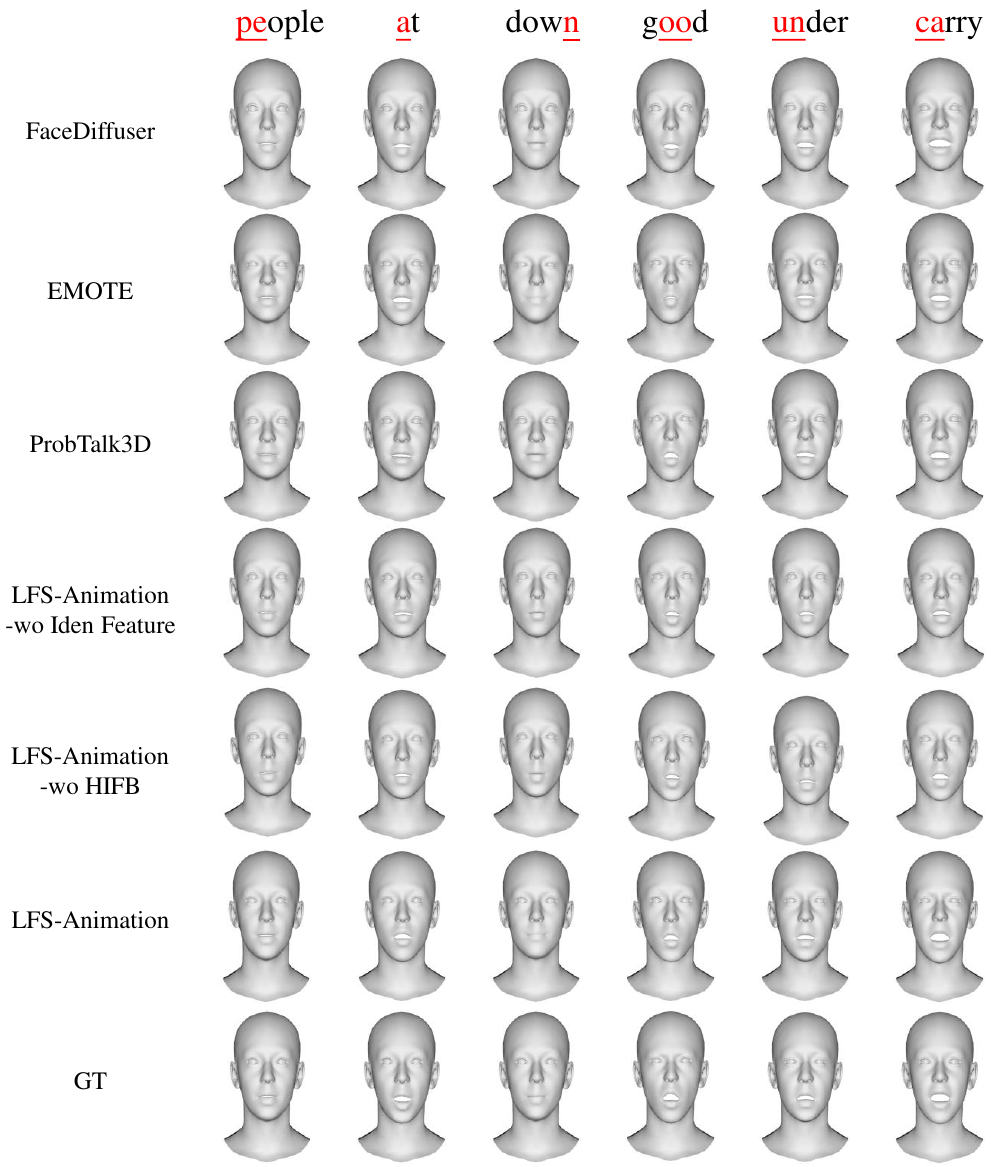}
  \caption{Visual comparison of generated facial animations by different models together with ground truth (GT).}
  \Description{}
  \label{fig:qualitative_lipsync}
\end{figure*}

\paragraph{Lip Synchronization.}
We evaluate lip motion quality on speech segments containing phonetically challenging structures such as bilabial consonants (/p/, as in \textit{“people”}), rounded vowels (\textit{“good”}), and open-mouth syllables (\textit{“under”, “carry”}). These audio sequences are selected from the 3DMEAD test set with neutral emotion. We compare our model \textit{LSF-Animation} with \textit{FaceDiffuser}, \textit{ProbTalk3D}, and the deterministic \textit{EMOTE} (using its official release). 
In addition, we also include two ablation variants for qualitative comparison: one without the identity feature (``w/o Iden Feature'') and another without the Hierarchical Interaction Fusion Block (``w/o HIFB'').
As shown in Fig.~\ref{fig:qualitative_lipsync}, \textit{LSF-Animation} generates more accurate and expressive lip movements, closely matching the ground truth, while other methods often miss fine-grained articulation. The ablation variants exhibit degraded performance, confirming the importance of both the identity feature and HIFB in producing natural and synchronized lip motions.

\begin{figure}
  \includegraphics[width=\linewidth]{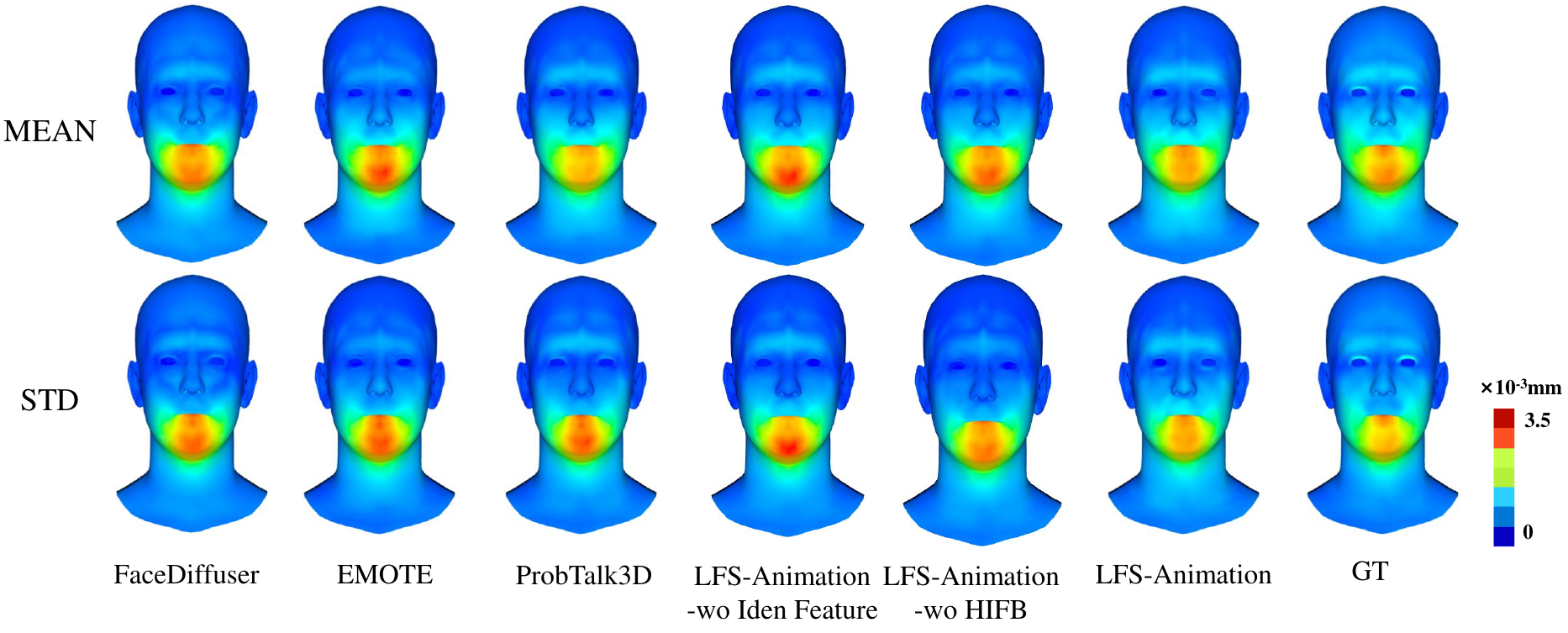}
  \caption{Comparison using heatmap visualization of mean and standard deviation of generated animation by different models together with ground truth (GT) given audio sequence, uttering the sentence: "How good is your endurance
". Our LFS-Animation closely replicates the dynamics of ground-truth sequence.}
  \Description{}
  \label{fig:heatmap}
\end{figure}

\paragraph{Motion Dynamics.}
Following~\cite{probtalk3d}, we visualize the mean and standard deviation of the vertex-wise L2 distances between adjacent frames within a sequence. A higher mean indicates more frequent motion (depicted as warmer colors), while a higher standard deviation reflects richer variation in facial dynamics. As shown in Fig.~\ref{fig:heatmap}, \textit{LSF-Animation} closely replicates the dynamics of ground-truth sequences, especially in the mouth and cheeks regions.

\begin{figure*}
  \includegraphics[width=0.8\textwidth]{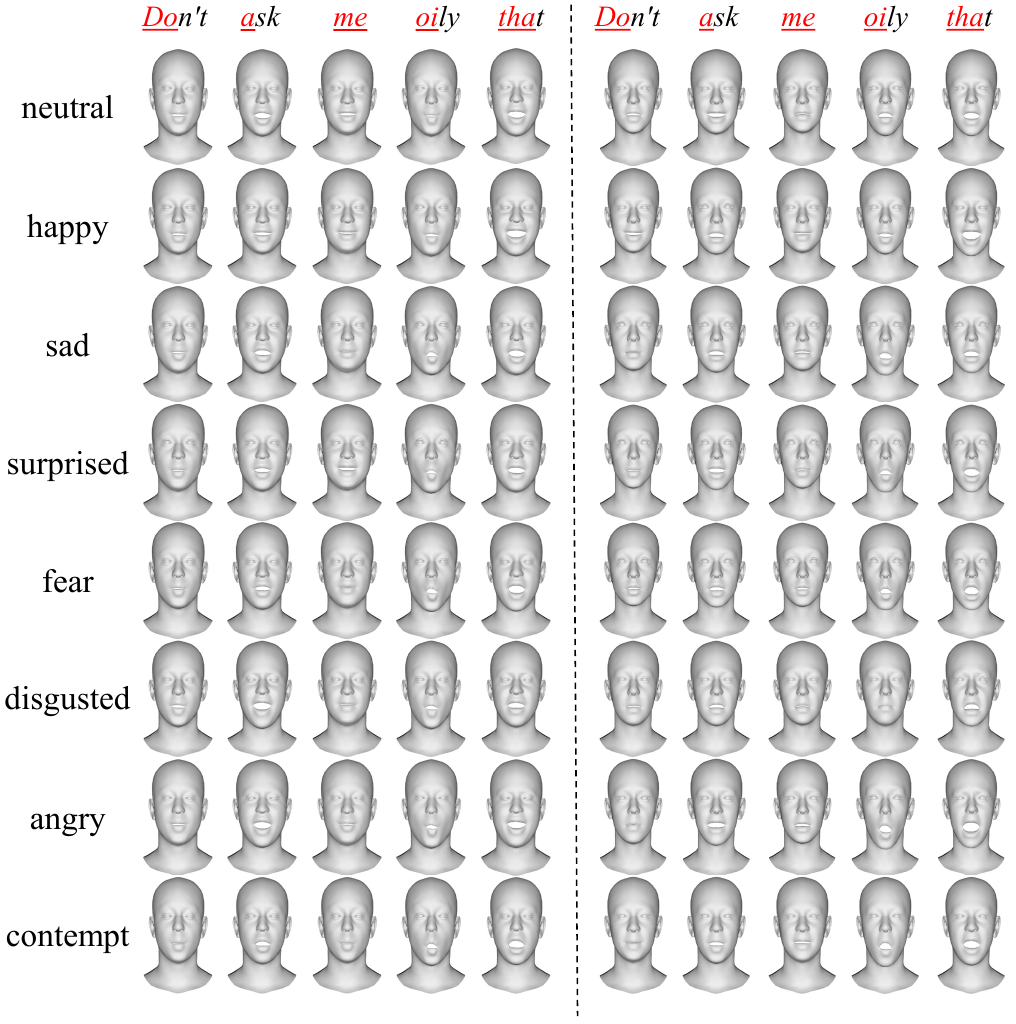}
  \caption{ Visual results of facial animations generated by LSF-Animation for the audio sequence: "Don't ask me to carry an oily rag like that". Each row corresponds to a different emotion, and two subjects are shown to illustrate the consistency of emotion control and identity generalization across varying emotional conditions.}

  \Description{.}
  \label{fig:emo_visualization}
\end{figure*}

\paragraph{Label-Free Representation Visualization.}
To demonstrate the expressiveness and generalization of our label-free representation, Fig.~\ref{fig:emo_visualization} shows facial animations generated using the same spoken sentence expressed under different emotional states and subject identities. The input consists of audio clips with varying emotions but identical phonetic content, allowing a direct comparison of emotional modulation. The clear differences across emotions and the consistent animation quality across unseen identities highlight our model's ability to generate diverse and identity-agnostic facial expressions without relying on explicit emotion or identity labels.

\subsection{Ablation Analysis}

Table~\ref{tab:ablation_results} presents the results of our ablation study, categorized into \textit{representation strategy} and \textit{fusion strategy}.

\paragraph{Representation Strategy Analysis.} 
We begin by assessing the role of emotion and identity conditioning in generating expressive and personalized facial animations. The ``w/o style one-hot'' variant, which only uses HuBERT-extracted motion features without any emotion or identity input, performs the worst across all metrics. This highlights the necessity of integrating external conditioning information beyond pure speech content. 
Furthermore, when no identity information is provided (``w/o iden''), the decoded expression parameters fail to generalize to different speakers, leading to the worst overall performance. Incorporating full shape parameters as identity features (``Emo one-hot + shape parameter'') yields some improvements but remains inferior to the baseline due to redundant information in the raw shape vectors.

To evaluate different representation formats, we compare implicit feature encodings with ground-truth labels. The ``Baseline'' setting uses one-hot labels for both emotion and identity, serving as an upper-bound reference. Our ``+ Emotion Feature'' variant replaces the ground-truth emotion label with frame-level emotion embeddings extractd from Emotion2vec. This variant achieves comparable performance to the baseline, validating that emotional cues can be effectively extracted from raw audio without explicit labeling.

To further enable a fully label-free facial animation pipeline and enhance generalization to unseen identities, we incorporate identity features captured from the neutral mesh. The ``+ Emotion Feature + Iden Feature'' variant replaces both ground-truth labels with style vectors constructed from these extracted emotion and identity features. This setting yields improvements over the baseline in multiple accuracy-based metrics: a 4.27\% reduction in MVE (from 1.2933 to 1.2380), a 9.47\% reduction in LVE (from 1.2708 to 1.1504), an 8.63\% reduction in MEE (from 1.1852 to 1.0828), and a 7.83\% reduction in CE (from 1.0691 to 0.9753). However, we observe a 0.25\% drop in FDD (from 0.4845 to 0.4867) and a 9.30\% drop in Diversity (from 0.4310 to 0.3909), which is likely due to the sequence-level nature of the style-vector-based approach, limiting the model's capacity to model fine-grained temporal variations. 
These results highlight both the effectiveness and current limitations of style-vector–based implicit representations under a fully label-free setting.

\paragraph{Fusion Strategy Analysis.} 
While the representation strategy incorporates emotion and identity information as a style vector, it only injects global information once at the beginning and fails to preserve frame-level emotional variation. To address this limitation, we explore a dual-branch architecture that integrates frame-level emotion features extracted from speech with motion features obtained from the HuBERT branch.
We compare three different fusion strategies: (1) a gate-based late fusion~\cite{lu2024fc}; (2) a cross-attention-based late fusion~\cite{faceformer}; and (3) our proposed \textit{HIFB}.
Both gate-based and cross-attention strategies are considered late fusion because they operate after each modality’s features have been independently encoded. In contrast, our HIFB allows modality-specific streams to co-evolve through multi-level interaction, better preserving emotional dynamics and temporal alignment.

All three fusion strategies yield consistent improvements over the style-vector-based approach in terms of overall motion modeling accuracy, as reflected by reductions in MVE, LVE, MEE, and CE. This confirms the effectiveness of leveraging frame-level emotional information during motion synthesis. However, the gate fusion and cross-attention fusion variants both exhibit significantly degraded FDD scores, indicating poor modeling of upper-face dynamics. This can be attributed to their late fusion design, which combines emotion and motion features only after independent encoding. Such delayed interaction may weaken the alignment between subtle high-frequency facial movements, particularly in the eyebrow and forehead regions, and corresponding emotional cues.

In contrast, our proposed HIFB enables progressive, bidirectional interactions throughout the encoding process, allowing the network to capture fine-grained emotional dynamics with greater temporal precision. It achieves substantial improvements in FDD, reducing it by 49.4\% compared to cross-attention fusion and 38.4\% compared to gate fusion. In addition, HIFB delivers comparable or superior results across all other metrics. Relative to the ``Baseline (gt one-hot) '' model, HIFB yields improvements of 5.3\% in MVE, 13.5\% in LVE, 2.5\% in FDD, 14.1\% in MEE, and 13.7\% in CE. These results demonstrate that our HIFB fusion mechanism effectively enhances both spatial fidelity and expressive temporal modeling in a fully label-free setting.

Overall, our fusion strategy analysis highlights the necessity of temporal-level emotion-motion alignment and the superiority of HIFB in achieving this goal.

\subsection{User Study}

To assess perceptual quality, we conduct an A/B test comparing our model with ground-truth animations and two strong SOTA methods, FaceDiffuser-DDIM and ProbTalk3D. For each model, 24 videos are generated across 8 emotions and 3 unseen identities using audio from 3DMEAD and in-the-wild sources. An additional 24 clips from our model are used for comparison with held-out ground-truth sequences.
Participants view 18 randomly sampled video pairs (one per emotion) and compare animations across three criteria: lip synchronization, facial realism, and emotional expressiveness. The survey is deployed via Wenjuanxing\footnote{\url{https://www.wjx.cn/}}, a widely used Chinese survey platform, and distributed among computer vision and graphics researchers.

\begin{figure*}[t]
  \includegraphics[width=\linewidth]{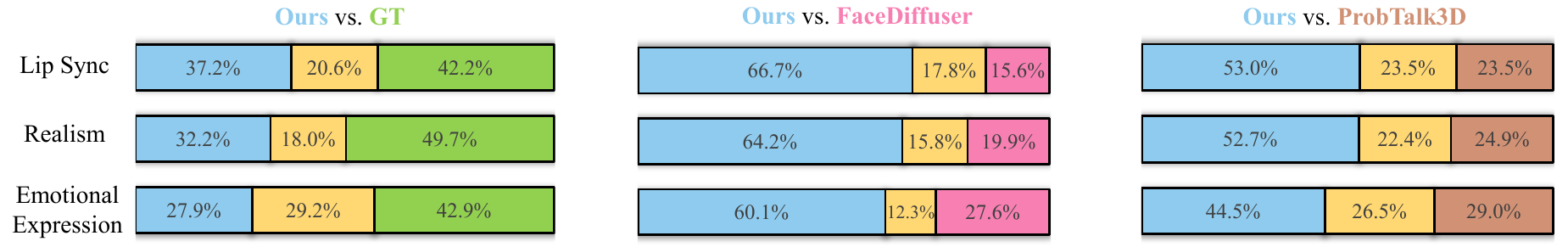}
  \caption{User preference comparison of LSF-Animation against ground truth (GT), FaceDiffuser, and ProbTalk3D. Colored bars represent the proportion of votes for each model, with yellow indicating a perceived tie in quality.
}
  \Description{}
  \label{fig:user_study}
\end{figure*}

We collected a total of 65 responses, with 4 discarded due to failing attention checks, resulting in 61 valid entries. As shown in Fig.~\ref{fig:user_study}, while our model is slightly less preferred than the ground truth, which is expected and consistent with prior work, it outperforms both FaceDiffuser and ProbTalk3D across all three evaluation criteria. These results highlight the effectiveness of our method in generating 3D facial animations with accurate lip-sync, high visual realism, and emotionally expressive quality.

\section{Conclusion}
In this paper, we propose LSF-Animation, a speech-driven 3D facial animation representation that removes the dependency on prior labels of emotion and identity by learning expressive representations directly from raw audio and a neutral face mesh. Combined with our HIFB, the model enables fine-grained integration of emotion and motion features, leading to enhanced upper-face expressiveness. Experiments on the 3DMEAD dataset demonstrate that our method achieves state-of-the-art performance in both accuracy and diversity compared to recent deterministic and probabilistic baselines.

While our method advances label-free personalized facial animation, future work will explore more comprehensive solutions for generating fully textured 3D talking heads in real-world open-domain scenarios.

\begin{acks}
This work was supported by the National Natural Science Foundation of China (U21A20515, 62476262, 62206263, 62271467, 62306297, 62306296, 62202076), the Beijing Natural Science Foundation (4242053, L242096), and the Fundamental Research Funds for the Central Universities. This research was also supported by Zhongguancun Academy, Beijing, China (Project No. 20240305).
\end{acks}

\clearpage

\clearpage

\bibliographystyle{ACM-Reference-Format}
\bibliography{sample-base}

\end{document}